\documentclass{article}
\usepackage[utf8]{inputenc}
\usepackage{spconf,amsmath,graphicx,amssymb,array,multirow,float,subcaption,makecell}
\usepackage{xcolor}
\floatstyle{plaintop}
\restylefloat{table}

\title{A Highly Adaptive Acoustic Model for Accurate Multi-Dialect Speech Recognition}
\name{Sanghyun Yoo$^{1}$ \qquad Inchul Song$^{1}$ \qquad Yoshua Bengio$^{2}$\sthanks{CIFAR Senior Fellow \& AI Chair.}}
\address{$^{1}$Samsung Advanced Institute of Technology, Republic of Korea \\
$^{2}$Mila, Université de Montréal, Canada \\
\{sam.yoo, inchul2.song\}@samsung.com \qquad yoshua.bengio@mila.quebec}
\usepackage{fancyhdr}
                    \fancypagestyle{plain}{
  \fancyhf{}
  \fancyhead{}
  \fancyfoot{}
}
\fancypagestyle{firstplain}{
  \fancyhf{}
  \fancyfoot[C]{\fontsize{8}{10} \selectfont © 20XX IEEE. Personal use of this material is permitted. Permission from IEEE must be obtained for all other uses, in any current or future media, including reprinting/republishing this material for advertising or promotional purposes, creating new collective works, for resale or redistribution to servers or lists, or reuse of any copyrighted component of this work in other works.}
  \fancyfoot[R]{}
}
\begin{document}

\maketitle\thispagestyle{firstplain}

\ninept
\maketitle
\begin{abstract}
Despite the success of deep learning in speech recognition, multi-dialect speech recognition remains a difficult problem. Although dialect-specific acoustic models are known to perform well in general, they are not easy to maintain when dialect-specific data is scarce and the number of dialects for each language is large. Therefore, a single unified acoustic model (AM) that generalizes well for many dialects has been in demand. In this paper, we propose a novel acoustic modeling technique for accurate multi-dialect speech recognition with a single AM. Our proposed AM is dynamically adapted based on both dialect information and its internal representation, which results in a highly adaptive AM for handling multiple dialects simultaneously. We also propose a simple but effective training method to deal with unseen dialects. The experimental results on large scale speech datasets show that the proposed AM outperforms all the previous ones, reducing word error rates (WERs) by 8.11\% relative compared to a single all-dialects AM and by 7.31\% relative compared to dialect-specific AMs.
\end{abstract}
\section{Introduction}
\label{sec:intro}
Every language has variations in terms of pronunciation. For example, English spoken by a native speaker is different from the one spoken by a non-native speaker, e.g., an Asian. Therefore, an acoustic model (AM) trained with speech data from only native speakers often fails to recognize speech from non-native speakers. An effective approach to dealing with multi-dialect speech recognition is to train a dialect-specific AM for each dialect \cite{melfeky}. Although it performs well in general, one disadvantage is that a separate AM needs to be maintained for each dialect, which increases operational cost. Thus, there have been demands for a single AM that can be used to recognize many dialects accurately.

A straightforward way to build a single AM for multiple dialects is to train the model on mixed data from many dialects. Although such an AM usually underperforms dialect-specific AMs, it can be used as a good starting point and hence is commonly considered as the baseline model in the literature \cite{melfeky,bli,mchen,tkim}. Recently, there have been many attempts to improve the single AM approach for multi-dialect speech recognition. One such approach is based on multi-task learning, where an AM is trained to not only predict phonemes but also identify dialects \cite{melfeky,bli,xyang,ajain}. There are other approaches that provide the AM with an auxiliary input such as i-vectors \cite{ndehak} or dialect information to make it adaptive to different dialects \cite{bli,mchen,ajain}.

Feature-wise Linear Modulation (FiLM) is a recently introduced adaptive neural network modeling technique \cite{eperez,hvries,fstrub,vdumoulin2}. By applying feature-wise transformations based on an auxiliary input, it enables neural networks to effectively adapt to multiple sources of information in a large set of problems \cite{vdumoulin2}. Although FiLM was originally proposed for a visual QA problem \cite{eperez,hvries,fstrub}, it is a general approach and hence can also be applied to acoustic modeling.

In this paper, we propose a novel approach to building a single AM for multi-dialect speech recognition. Motivated by FiLM, we apply feature-wise transformations to the AM based on dialect information. However, unlike the original FiLM architecture that adapts neural networks based on only external input, we also use internal representation extracted within the network as additional conditioning information. Using the combination of both external and internal information in feature-wise transformations makes our AM more adaptive and able to deal with multiple dialects more effectively. We also propose a simple but effective training method to handle unseen dialects during training. Through experimental evaluation on large scale speech datasets, we show that our proposed AM outperforms all the previous ones in terms of WERs.

The rest of this paper is organized as follows. First, Section \ref{sec:related_work} discusses past research related to this work. Section \ref{sec:proposed_method} describes our proposed AM that is adapted by both dialect information and internal representation. Section \ref{sec:experimental_results} shows the experimental results, and Section \ref{sec:conclusion} concludes the paper.

\section{Related Work}
\label{sec:related_work}
The traditional approach to dealing with multi-dialect speech is to build a different AM per dialect as if each dialect is a different language \cite{melfeky}. However, when dialect-specific data is scarce, such a dialect-specific AM underperforms even a single AM trained with data from all dialects \cite{xyang}. This is because a dialect-specific AM is not able to learn similarities between different dialects. To deal with such a resource-scarcity problem, there are some approaches that jointly train dialect-independent parts of an AM with data from all dialects and separately train dialect-dependent parts with dialect-specific data \cite{mchen, xyang, yhuang}. In other approaches, dialect-specific AMs are obtained simply by training a single AM using all available data first and then fine-tuning it on dialect-specific data \cite{bli}. However, the main drawback of these methods is that they all need to maintain several models for a given language, which incurs high maintenance cost especially when speech recognition service is provided for many languages and dialects \cite{melfeky}.

A single AM for all dialects has merit in terms of maintenance cost, and hence there have been many attempts to improve its performance. Some of them are based on multi-task learning \cite{melfeky,bli,xyang,ajain} where an AM is trained to predict phonemes as well as classify dialects. However, it is not easy to train such a model because deciding the weights given to two different tasks is not straightforward \cite{xyang}. Other approaches use an auxiliary input such as i-vectors or dialect information \cite{bli,mchen,ajain}. \cite{bli} proposes to provide dialect information as an additional input of an end-to-end speech recognition model, and \cite{ajain} proposes to use utterance-level dialect embedding extracted from a dialect classifier. It has been shown that they outperform dialect-specific models \cite{bli}. 

Conditioning is an effective method for adaptive neural network modeling. \cite{vdumoulin} shows that conditional instance normalization, which dynamically generates the scaling and shifting parameters in instance normalization, can transfer styles of images succinctly. Motivated by this work, \cite{tkim} proposes to dynamically generate the parameters in layer normalization \cite{jba} for speaker adaptation. Later, a general-purpose conditioning approach called Feature-wise Linear Modulation (FiLM) is proposed \cite{eperez,hvries,fstrub,vdumoulin2}. In this approach, special layers called FiLM layers are inserted into the network, and each such layer applies feature-wise affine transformations to its input as follows:
\[\hat{\mathbf{x}} = \mathbf{\gamma} \odot \mathbf{x} + \mathbf{\beta}\]
where $\odot$ indicates pointwise product of vectors, $x$ is a FiLM layer's input vector, and $\gamma$ and $\beta$ are the scaling and shifting vectors that are dynamically generated based on an auxiliary input. Although FiLM is proposed to solve a visual QA problem, such feature-wise transformations can also be applied to solve a diverse set of problems \cite{vdumoulin2}. However, to the best of our knowledge, there has been no work that uses FiLM for multi-dialect speech recognition. In this paper, we examine how FiLM can be applied to acoustic modeling for accurate multi-dialect speech recognition and propose an improved network architecture.

\section{Multi-Dialect Acoustic Modeling}
\label{sec:proposed_method}
\subsection{Baseline Acoustic Models}
\label{ssec:baseline_arch}
All AMs considered in this paper use a uni-directional recurrent neural network with LSTM layers followed by a softmax layer. We apply batch normalization to the input of each LSTM layer as in \cite{claurent} and also add a lookahead convolutional layer \cite{damodei} after each LSTM layer to provide the model with some future context. We call this basic AM that does not take any dialect information a \textit{dialect-unaware AM}. Then, we obtain \textit{dialect-specific AM}s by fine-tuning the dialect-unaware AM with data from the corresponding dialect. Our third baseline AM is to take dialect information as an additional input \cite{bli,ajain,ohamid}. As in \cite{bli}, dialect information is represented as a one-hot vector and concatenated with the other input features\footnote{Although the dialect information can be added to the input of every layer, there is not much difference in performance in our experiments.}. We call this model a \textit{dialect-aware AM}.

\begin{figure}
    \centering
    \begin{subfigure}[t]{1\columnwidth}
    \centering
        \includegraphics[height=1.05in]{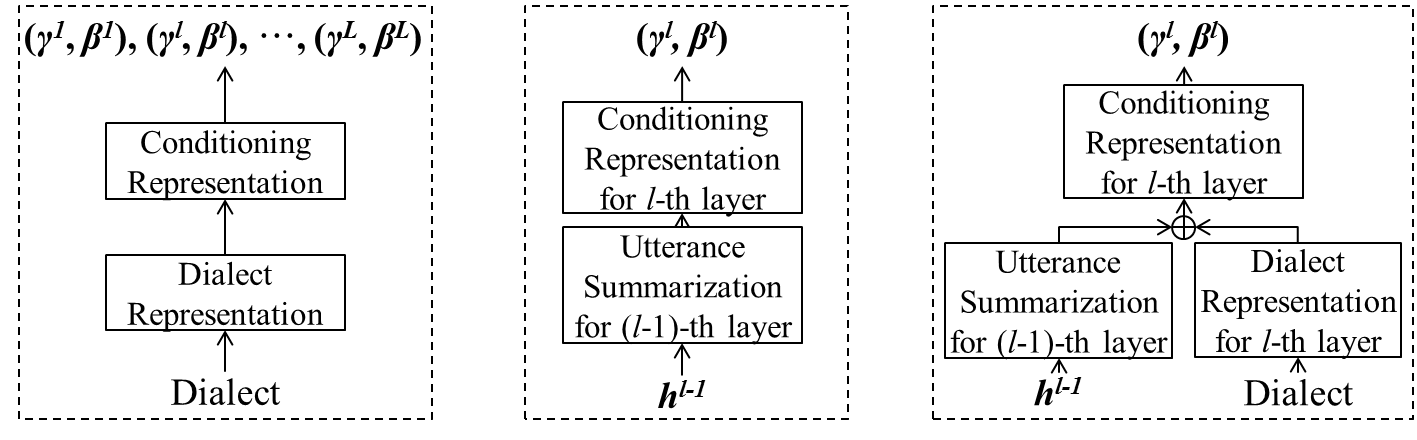}
        \caption{The $\gamma$ and $\beta$ parameters generated based on dialect information (left), utterance summarization (center), or both (right).}
        \label{figure:conditioning_info}
    \end{subfigure}
    \begin{subfigure}[t]{1\columnwidth}
    \centering
        \includegraphics[height=0.9in]{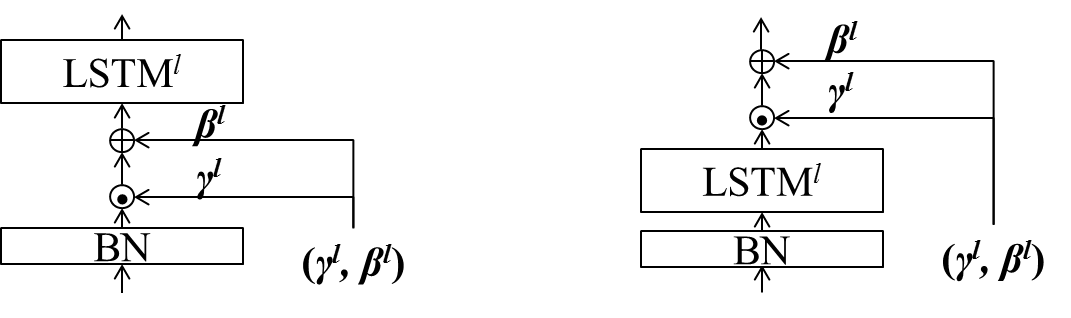}
        \caption{Conditioning applied to each layer's input (left) or output (right).}
        \label{figure:conditioning_positions}
    \end{subfigure}
    \caption{Illustration of (a) three kinds of conditioning information and (b) two different conditioning positions.}
    \label{fig:final_am}
\end{figure}

\subsection{Conditioning with External Information}
\label{ssec:conditioning_with_external_information}
The FiLM architecture can be applied to acoustic modeling for multiple dialects by using dialect information as a conditioning input. Based on this conditioning information, the scaling and shifting parameters for feature-wise affine transformations, $\gamma$ and $\beta$, are dynamically generated. If dialect information $d$ is represented as a $D$-dimensional one-hot vector where $D$ is the number of dialects, the $\gamma$ and $\beta$ for all layers can be generated at once as follows:
\[a_c = tanh(W_c(tanh(W_d d + b_d)) + b_c)\]
\[(\gamma^1, ... \gamma^L)=tanh(W_{\gamma} a_c + b_{\gamma}),
\quad (\beta^1, ... \beta^L)=tanh(W_{\beta} a_c + b_{\beta}),\]
where $L$ is total number of LSTM layers, $W$s are weight matrices, and $b$'s are bias terms. This is shown  on the left picture in Fig. \ref{figure:conditioning_info}.

The generated $\gamma$ and $\beta$ parameters are applied to either the input of each LSTM layer after batch-normalization (as shown on the left picture in Fig. \ref{figure:conditioning_positions}) or the output of each layer after activation (as shown on the right picture in Fig. \ref{figure:conditioning_positions}).

\subsection{Conditioning with Internal Representation}
\label{ssec:conditioning_with_internal_representation}
\cite{tkim} shows that an AM dynamically adapted based on the hidden representation of each layer performs well in speaker adaptation. Their method can also be applied to multi-dialect speech recognition. Similar to \cite{tkim}, we attach a separate utterance-level feature extractor network to each layer and jointly train it with the main AM. The feature vector $a_s^{l-1}$, called utterance summarization of $(l-1)$-th layer, is extracted as follows:
\[a_s^{l-1}=\frac{1}{T}\sum_{t=1}^{T}{tanh({W_s^{l-1}}{h_t^{l-1}}+b_s^{l-1})},\]
where $T$ is the number of total frames in the utterance, and $h_t^{l-1}$ is the output of the $(l-1)$-th hidden layer at time step $t$ ($h_t^0$ is the input features $x_t$). $W_s^{l-1}$ and $b_s^{l-1}$ are the weight matrix and bias term, respectively. We apply one more non-linear transformation before parameter generation. Therefore, the $\gamma^l$ and $\beta^l$ at layer $l$ are generated as follows:
\[a_c^l=tanh({W_c^l}{a_s^{l-1}}+b_c^l)\]
\[\gamma^l=W_{\gamma}^l a_c^l + b_{\gamma}^l, \quad\quad \beta^l=W_{\beta}^l a_c^l + b_{\beta}^l,\]
where $W^l$s are weight matrices, and $b^l$s are bias terms. This method is depicted on the center picture in Fig. \ref{figure:conditioning_info}.

\subsection{Conditioning with External and Internal Information}
\label{ssec:conditioning_with_external_and_internal_information}
Finally, we propose to use not only dialect information and but also utterance summarization in parameter generation. Conditioned by both external input and internal representation at the same time, an AM can be more adaptive in dealing with multiple dialects simultaneously than models described above.

First, a one-hot vector indicating the dialect is transformed into a hidden representation $a_d^l$ and concatenated with utterance summarization of the previous layer. Then, it is fed into another non-linear layer to make the final conditioning representation $a_c^l$ as follows:
\[a_d^l=tanh(W_{d}^l d + b_{d}^l)\]
\[a_c^l=tanh({W_c^l}{(a_d^l \oplus a_s^{l-1})} + b_c^l),\]
where $d$ is a one-hot vector representing the dialect, $a_s^{l-1}$ is the utterance summarization of $(l-1)$-th layer, which is introduced in Section \ref{ssec:conditioning_with_internal_representation}. The symbol $\oplus$ indicates the concatenation operation. The $\gamma^l$ and $\beta^l$ at $l$-th layer are generated based on this conditioning representation $a_c^l$ as shown on the right picture in Fig. \ref{figure:conditioning_info}:
\[\gamma^l=W_{\gamma}^l a_c^l + b_{\gamma}^l, \quad\quad \beta^l=W_{\beta}^l a_c^l + b_{\beta}^l.\]
As before, $W$s and $b$'s indicate the weight matrices and bias terms.

\subsection{Unseen Dialect Handling}
\label{ssec:unseen_dialect_handling}
One limitation of our model is that users always need to provide dialect information. Therefore, even for users speaking unseen dialects during training, one dialect (usually native) must be chosen, which may result in performance degradation. Using a dialect classifier might be one solution \cite{xyang,ajain}, but it makes the overall performance largely dependent on the classification accuracy. In this paper, we handle the unseen dialect problem by simply adding one more dialect, called $unknown$. During training, utterances randomly sampled with a certain probability (e.g., 0.1) are treated as if they were from the unknown dialect. At test times, users speaking unseen dialects are considered to use the unknown dialect and conditioning representation is generated accordingly.

\begin{table}[t]
  \centering
  \caption{The statistics of various datasets in terms of the number of utterances. (The numbers in parentheses refer to hours.)}
  \label{table:corpus_statistics}
  \begin{tabular}{|c|r||c|r|}
    \hline
    \textbf{Speaker} & \textbf{Training set} & \textbf{Speaker} & \textbf{Training set} \\ \hline
    Native     & 281,241 (960)   &
    Chinese         & 73,413 (131)     \\ \hline
    German          & 55,836 (74)        &
    Spanish         & 75,762 (102)                    \\ \hline
    French          & 47,881 (59)        &
    Italian         & 43,863 (57)       \\ \hline
    Portuguese      & 35,933 (46)       &
    Korean      & N/A (N/A)            \\ \hline
  \end{tabular}
\end{table}

\setlength{\tabcolsep}{0.57em}
\begin{table*}[htbp]
  \begin{tabular}{|l|c|r|r|r|r|r|r|r|r|r|r|r|r|}
    \hline
    \multicolumn{3}{|c|}{\textbf{Models}} &
    \multicolumn{10}{c|}{\textbf{WER}} \\ \cline{1-13}
    \multicolumn{1}{|c|}{\multirow{2}{*}{\textbf{Description}}} & \multicolumn{1}{c|}{\multirow{2}{*}{\makecell[c]{\textbf{Cond.}\\\textbf{Pos.}}}} &
    \multicolumn{1}{c|}{\multirow{2}{*}{\makecell[c]{\textbf{Size}\\\textbf{(M)}}}} & 
    \multicolumn{1}{c|}{\multirow{2}{*}{\textbf{Nat}}}    & \multicolumn{1}{c|}{\multirow{2}{*}{\textbf{CHI}}}    & \multicolumn{1}{c|}{\multirow{2}{*}{\textbf{GER}}}     & \multicolumn{1}{c|}{\multirow{2}{*}{\textbf{ESP}}}   & \multicolumn{1}{c|}{\multirow{2}{*}{\textbf{FRN}}}     & \multicolumn{1}{c|}{\multirow{2}{*}{\textbf{ITA}}}    & \multicolumn{1}{c|}{\multirow{2}{*}{\textbf{POR}}} & \multicolumn{1}{c|}{\multirow{2}{*}{\textbf{KOR}}} &
    \multicolumn{2}{c|}{\textbf{Overall}} \\ \cline{12-13}
    &&&&&&&&&&&\multicolumn{1}{c|}{\textbf{-KOR}}   & \multicolumn{1}{c|}{\textbf{+KOR}}                            \\ \hline\hline
    [M1] Dialect-unaware & \multirow{3}{*}{N/A} & 15.44 & 5.93 & 29.40 & 19.36 & 21.54 & 17.96 & 15.47 & 15.96 & 30.30 & 15.87 & 17.39 \\ \cline{1-1}\cline{3-13}
    [M2] Dialect-specific & & 15.44 & \textbf{5.79} & 28.32 & 18.00 & 20.36 & 16.98 & 14.86 & 15.16 & 33.20 & 15.12 & 17.24 \\ \cline{1-1}\cline{3-13}
    \makecell[l]{[M3] Dialect-aware} & & 15.46 & 5.95 & 28.51 & 18.51 & 20.55 & 17.33 & 14.96 & 15.17 & 37.70 & 15.33 & 17.69 \\ \hline\hline
    
    [M4] Cond. on D-Info & \multirow{3}{*}{Input} & 16.76 & 5.87 & 26.68 & 17.95 & 20.06 & \textbf{16.54} & 14.60 & 14.91 & 38.14 & 14.80 & 17.26 \\ \cline{1-1}\cline{3-13}
    \makecell[l]{[M5] Cond. on Utt-Sum} & &  16.90 & 5.96 & 27.69 & 18.06 & 20.46 & 17.87 & 15.10 & 15.38 & 29.04 & 15.28 & 16.73 \\ \cline{1-1}\cline{3-13}
    \makecell[l]{[M6] Cond. on Both} & & 16.84 & 5.91 & 26.31 & 17.39 & 19.63 & 16.63 & 14.55 & \textbf{14.27} & 36.79 & 14.58 & 16.92 \\ \hline\hline
    \makecell[l]{[M7] Cond. on D-Info} & \multirow{4}{*}{Output} & 15.78 & 6.00 & 27.78 & 18.61 & 20.54 & 16.93 & 15.00 & 15.24 & 37.01 & 15.24 & 17.54 \\ \cline{1-1}\cline{3-13}
    \makecell[l]{[M8] Cond. on Utt-Sum} & & 15.92 & 5.92 & 27.27 & 17.89 & 20.13 & 17.46 & 14.68 & 15.30 & 29.22 & 15.05 & 16.54 \\ \cline{1-1}\cline{3-13}
    \makecell[l]{[M9] Cond. on Both} & & 15.86 & 5.87 & \textbf{26.14} & 17.12 & \textbf{19.37} & 16.65 & \textbf{14.07} & 14.37 & 37.96 & \textbf{14.44} & 16.92 \\ \cline{1-1}\cline{3-13}
    \makecell[l]{[M10] Cond. on Both + Unk-D} & & 15.86 & 5.80 & 26.61 & \textbf{16.76} & 19.48 & 16.92 & 14.31 & 14.36 & \textbf{28.51} & 14.51 & \textbf{15.98}\\ \hline
  \end{tabular}
  \caption{The performance comparison among 10 different AMs in terms of WERs (\%) for Native, Chinese (CHI), German (GER), Spanish (ESP), French (FRN), Italian (ITA), Portuguese (POR), and Korean (KOR) Ensligh. Since KOR is treated as the unseen dialect during training, there are two different overall WERs, -KOR and +KOR, which indicate that KOR is removed from or included in test cases, respectively.}
  \label{table:experimental_results}
\end{table*}

\begin{figure}[htbp]
	\begin{subfigure}[t]{0.5\textwidth}
    	\centering
        \includegraphics[height=1.4in]{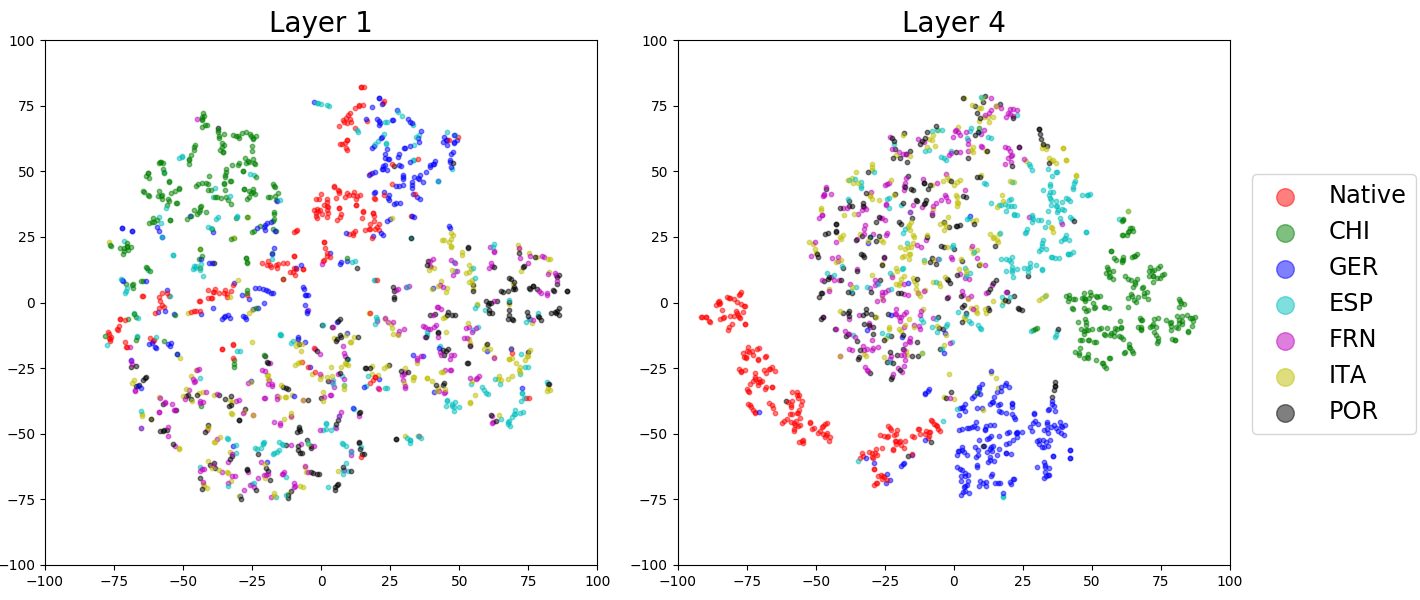}
        \caption{Using utterance summarization only.}
        \label{figure:tsne_utt_summarization}
	\end{subfigure}
	
    \begin{subfigure}[t]{0.5\textwidth}
    	\centering
        \includegraphics[height=1.4in]{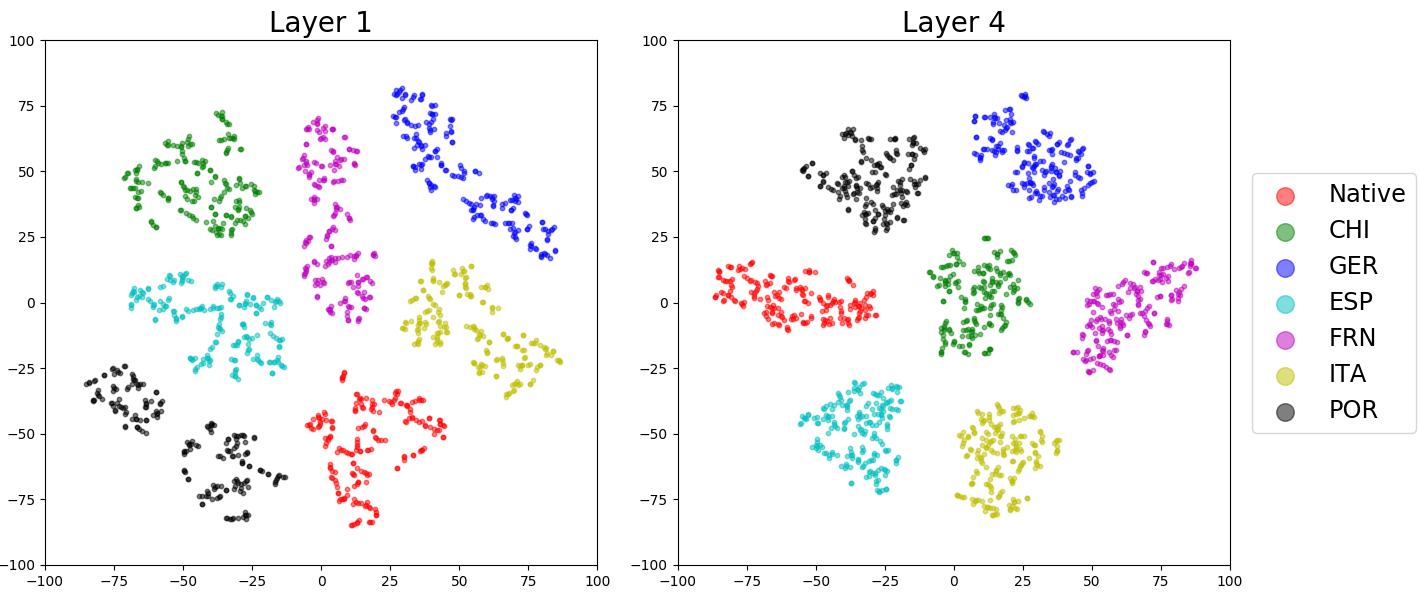}
        \caption{Using both dialect information and utterance summarization.}
        \label{figure:tsne_dialect_info_and_utt_summarization}
	\end{subfigure}%
    \caption{$\gamma$ and $\beta$ parameters for the input of the first layer (left) and the last layer (right) generated based on the corresponding information, plotted using t-SNE in 2D-space. Each color represents each dialect.}
\end{figure}

\section{Experimental Results}
\label{sec:experimental_results}
\subsection{Datasets}
\label{ssec:datasets}
We evaluate the models described above on large vocabulary speech recognition datasets, which are English speeches spoken by users of 8 different dialects including native, Chinese, German, Spanish, French, Italian, Portuguese, and Korean. For the native speech data, we use 960 hours of speeches from the LibriSpeech corpus \cite{vpanayotov}. For the nonnative speech data, we use the commercial datasets from Speech Ocean \cite{speechocean}. Among these, the Korean English dataset is not used during training for the ablation test. The statistics of each dataset are shown in Table \ref{table:corpus_statistics}. Note that nonnative datasets are much smaller than the native one, which is a common setting in multi-dialect speech recognition.

In all experiments, we use 80-dimensional log-mel features, computed with a 25ms window and shifted every 10ms. We follow the standard Kaldi recipe s5 \cite{dpovey} for preparing data.

\subsection{Network Architectures}
\label{ssec:network_architecture}
We evaluate 10 different architectures in our experiments as shown in Table \ref{table:experimental_results}. All AMs considered in the experiments have 4 uni-directional LSTM layers, each of which has 640 LSTM units. During training, the mini-batch size is set to 320, and the Adam optimizer \cite{dkingma} is used with the initial learning rate 0.1 and learning rate scheduling is applied.

M1 - M3 are the baseline models. M1 is the dialect-unaware model, which is trained with data from all dialects except Korean English. M2 indicates the dialect-specific models, i.e., seven models trained by fine-tuning M1 on data from corresponding dialects. M3 is the dialect-aware model that takes dialect information as an additional input.

M4 - M10 are AMs conditioned on dialect information (D-Info), utterance summarization (Utt-Sum), or both (Both). As indicated by the second column of Table \ref{table:experimental_results}, M4 - M6 apply feature-wise transformations to the input of each layer, and M7 - M10 apply them to the output. Although these two groups take the same information, M7 - M10 generate a smaller number of $\gamma$ and $\beta$ parameters than M4 - M6\footnote{The number of parameters for the input is four times larger than that for the output because an LSTM unit has three gates and one cell, and each one has an input.}, and hence the model sizes are reduced as shown on the third column in the table.

M4 and M7 are conditioned on only dialect information and use 64 hidden units for dialect representation. Similarly, M5 and M8 are conditioned on only utterance summarization and also use  64 hidden units for dialect representation. M6, M9, and M10, conditioned on both information, use 32 hidden units for each representation and then concatenate them together. In all cases, another fully-connected layer with 64 hidden units is located on top of the representation to make the final conditioning representation (the upper boxes in Fig. \ref{figure:conditioning_info}). M10 is the model using our unseen dialect handling technique (Unk-D) described in Section \ref{ssec:unseen_dialect_handling}\footnote{We assume that all models except M10 treat Korean English as the native one due to the reason described in Section \ref{ssec:unseen_dialect_handling}.}. With a probability of 0.1, this model treats an utterance as the one coming from the unknown dialect.

\subsection{Results}
\label{ssec:results}
The Word Error Rates (WERs) of different AMs over various dialects and overall WERs are shown on the corresponding columns in Table \ref{table:experimental_results}. As we do not use Korean English at training time for the ablation test, the overall WERs are differently calculated according to whether Korean English is removed from or included in test cases.

M1 is our baseline and its WERs for other dialects are much higher than that for the native one, so we can see that the characteristics of the native speech are different from those of the other dialects. The performance of M2 is improved for all dialects except for Korean English, which is handled by the native English model that is more tuned to the native English than M1. Feeding dialect information as input vectors seems effective, as also shown in \cite{bli}, because M3 outperforms M1 and is competitive with M2.

M4 is the model that applies feature-wise transformations to the input of each layer based on only dialect information. This model outperforms even M2 with a 6.74\% relative WER reduction, which indicates that feature-wise transformations are effective.

M5 is the AM adapted by only utterance summarization, which is similar to the model proposed in \cite{tkim}. Although it underperforms M4, its performance is competitive with M2 or M3. Moreover, its performance for the unseen dialect case is better than that of M1. This means that internal representation plays an important role to make an AM adaptive to multiple dialects. To show how different the $\gamma$ and $\beta$ parameters are depending on the dialect, we visualize them by using t-SNE \cite{lvdmaaten} in Fig. \ref{figure:tsne_utt_summarization}. Although no dialect information is given to the model, the parameters tend to be clustered according to the dialect, especially in the higher layers. This shows that conditioning information based on utterance summarization is able to catch dialect-specific aspects well.

M6 is the AM conditioned on both dialect information and utterance summarization. The overall WER is reduced by 3.57\% relative compared to M2 when Korean English test cases are not considered. When we compare M6 with the single baseline AM, M3, the overall WER is reduced by 4.89\% relative. Fig. \ref{figure:tsne_dialect_info_and_utt_summarization} shows the visualization of $\gamma$ and $\beta$ parameters generated in M6. We can see that they are more strongly clustered by dialects, compared to those from M5. Similar to M5, the parameters in the higher layers seem to be more related to dialect-specific information.

M7 - M9 are equivalent to M4 - M6, respectively, except that these models apply feature-wise transformations to the output of each layer. Both of M7 and M8 show competitive performance with M2 or M3. Unlike M4, using only dialect information does not obtain better performance than the baseline models. However, M9, which uses the combination of dialect information and utterance summarization is superior to all the other AMs in terms of the overall WER. Compared to M2, the WER is reduced from 15.12\% to 14.44\% (4.50\% relative). Moreover, it even outperforms M6 despite its smaller model size, which shows that the output of the layer is a better place to apply conditional transformations.

Finally, M10 is M9 with the unseen dialect handling technique applied. Although M10 performs similarly as M9 for all known dialects, it performs the best for Korean English, which is unseen during training. If Korean English is considered, the overall WER is reduced from 17.24\% to 15.98\% (7.31\% relative), compared to M2. As a single AM for all dialects, the overall WER is reduced by 8.11\% relative and 9.67\% relative, compared to M1 and M3, respectively.

\section{Conclusion}
\label{sec:conclusion}
In this paper, we propose a highly adaptive AM for accurate multi-dialect speech recognition. Unlike previous work, our AM is conditioned on both dialect information and internal representation. We also propose a simple but effective method for handling unseen dialects. We show through experimental evaluation that our proposed AM is a single model with low maintenance cost that outperforms all the previous AMs for multi-dialect speech recognition.

As future work, we plan to apply our proposed highly adaptive acoustic modeling technique to end-to-end speech recognition. Since end-to-end speech recognition models contain both the acoustic and language models, we expect that each part can benefit from our proposed technique.

\section{Acknowledgements}
\label{sec:acknowledgements}
YB would like to thank NSERC and CIFAR AI Chairs for funding.

\bibliographystyle{IEEEbib}
\bibliography{strings,refs}

\end{document}